\documentclass[10pt]{article}
\usepackage[utf8]{inputenc}
\usepackage[T1]{fontenc}
\usepackage{amsmath}
\usepackage{amsfonts}
\usepackage{amssymb}
\usepackage[version=4]{mhchem}
\usepackage{stmaryrd}

\title{PLN and NARS \\ Often Yield Similar  {\it strength $\times$ confidence} \\ Given Highly Uncertain \\ Term Probabilities }

\author{Ben Goertzel \footnote{SingularityNET, TrueAGI}}
\date{}

\begin{document}
\maketitle

\begin{abstract}
We provide a comparative analysis of the deduction, induction, and abduction formulas used in Probabilistic Logic Networks (PLN) and the Non-Axiomatic Reasoning System (NARS), two uncertain reasoning frameworks aimed at AGI.   

One difference between the two systems is that, at the level of individual inference rules, PLN directly leverages both term and relationship probabilities, whereas NARS only leverages relationship frequencies and has no simple analogue of term probabilities.   Thus we focus here on scenarios where there is high uncertainty about term probabilities, and explore how this uncertainty influences the comparative inferential conclusions of the two systems.  We compare the product of strength and confidence ( $s \times c$ ) in PLN against the product of frequency and confidence ( $f \times c$ ) in NARS (quantities we refer to as measuring the "power" of an uncertain statement) in cases of high term probability uncertainty, using heuristic analyses and elementary numerical computations.  

We find that in many practical situations with high term probability uncertainty, PLN and NARS formulas give  very similar results for the power of an inference conclusion, even though they sometimes come to these similar numbers in quite different ways.  
\end{abstract}

\section{Introduction}
Reasoning under uncertainty is a central challenge in artificial intelligence, and arguably critical to achieving AGI.  However, standard methodologies from the AI literature meet this challenge only partially.  Classical probabilistic and statistical reasoning frameworks like Bayes Nets and Markov Logic Networks don't effectively encompass advanced logical knowledge representation features like quantifiers or higher order expressions.  Statistical approaches like LLMs fail at logical reasoning in glaring and well-documented ways, in spite of their often remarkable inferential achievements (Goertzel, 2023). 

Two prominent frameworks aiming to address these shortcomings are:

\begin{enumerate}
  \item Probabilistic Logic Networks (PLN), developed by Ben Goertzel and colleagues, which combines probabilistic reasoning with a mix of higher-order predicate logic and term logic, driven by an observation-grounded semantics (Goertzel, 2009)
  \item Non-Axiomatic Reasoning System (NARS), developed by Pei Wang, which provides an uncertain term logic for evidence-based reasoning without relying on probabilistic assumptions. (Wang, 1994)
\end{enumerate}

Both of these systems provide inference rules for deduction, induction, and abduction, and then contain systems for extrapolating these basic rules to handle more advanced inference scenarios involving quantifiers, causal and temporal reasoning, and so forth.  Further, the two systems have in common that they assign both {\it strength} and {\it confidence} values to logical statements and relationships.  

However, the two systems differ fundamentally in their semantics of uncertainty and their use of term probabilities.  As one important aspect of this foundational semantic difference, PLN assigns probabilities to both terms and relationships, whereas NARS assigns uncertainties only  to relationships but not terms. 

This paper compares the basic truth value formulas of PLN and NARS (across the deduction, induction and abduction inference rules), focusing in particular on the scenario where there is high uncertainty about term probabilities in PLN.   Our main result is if one compares the product {\it strength $\times$ confidence} (which we call the ''power'') across PLN and NARS, then in situations of high term probability uncertainty, we find the two systems often give very nearly the same evaluation of the power to be associated to the result of an inference.

What is interesting is that the two systems often come to almost the same number but by different routes: i.e. the strong alignment we observe in the high-term-probability-uncertainty case does not occur if one looks at {\it strength} along or {\it confidence} alone, only if one looks at the product.

No similar, inference-rule-level alignment between the two systems occurs in the case of higher term probability confidence, as in these cases PLN's inductive and abductive conclusions become strongly Bayesian, which is a quite different direction from NARS.
 
\section{Background}
\subsection{Non-Axiomatic Reasoning System (NARS)}

The cognitive and philosophical theory underlying NARS is quite deep and involved and we will not attempt to summarize it here, referring the reader to (Wang, 2006).   Operationally, the core of the theory includes a set of term logic inference rules, and then uncertain truth value formulas mapping the uncertainties of the premises into the uncertainty of the conclusion.   Uncertainties are represented using $(f,c)$ {\it = (frequency, confidence) } pairs with a particular semantics.  The truth value formulas are obtained leveraging certain axiomatic assumptions that are adopted based on conceptual considerations.  The rules and formulas are viewed as allowing a reasoning system to adaptively generalize from experiential evidence using limited resources without making unwarranted assumptions such as prior probability distributions.

The rough semantics of the uncertain truth value components is,

\begin{itemize}
  \item Frequency $(f)$ : The proportion of positive evidence supporting a statement.
  \item Confidence (c): Reflects the amount of evidence relative to an experience parameter $k$, calculated via
  
  $$
  c = \frac{n}{n+k}
  $$
  \noindent where $n$ is the number of evidential observations supporting the statement
\end{itemize}

NARS also include the concept of

\begin{itemize}
\item Expectation $= f * c$
\end{itemize}

\noindent This quantity occurs here and there in work with NARS, where it is considered as an estimation of the overall reliability or utility of a statement in terms of how frequently it is observed (frequency) and how much evidence there is supporting it (confidence).  It is {\it not} conceptually the same or closely related to the probabilistic concept of expectation (meaning expected value).   

For purposes of comparison of NARS with PLN, the term "expectation" here is confusing and we will use instead the term

\begin{itemize}
\item Power $= f * c$
\end{itemize}

\noindent with a similar meaning for PLN (where "expectation" has its usual probabilistic meaning).   This quantity is important here because our core finding is that under assumptions of highly uncertain term probabilities, PLN and NARS often associate rather similar power values to the conclusions of inferences.   

We now summarize the quantitative truth value formulas associated with the three "first order term logic" rules of NARS logic.  NARS contains a number of other inference rules handling matters such as quantifiers, temporal and causal reasoning, but we will restrict attention here to these three basic rules, as they exemplify the core conceptual issues behind NARS truth value formulas and the crux of the similarities and differences with PLN.

\paragraph{ Deduction}. Term logic deduction is basically transitive inference.   Given:
  
\begin{itemize}
  \item $A \rightarrow B$ with $\left(f_{1}, c_{1}\right)$
  \item $B \rightarrow C$ with $\left(f_{2}, c_{2}\right)$
  \end{itemize}
  
\noindent one infers

\begin{itemize}
\item $A \rightarrow C$ with $(f, c)$
\end{itemize}

\noindent  obtaining the conclusion truth value via 

\begin{itemize}
  \item Frequency:

$$
f=\frac{f_{1} f_{2}}{f_{1}+f_{2}-f_{1} f_{2}}
$$

\item Confidence:

$$
c=c_{1} c_{2}\left(f_{1}+f_{2}-f_{1} f_{2}\right)
$$

\item Power:  $f \times c$ :

$$
f \times c=f_{1} f_{2} c_{1} c_{2}
$$
\end{itemize}

Note that here the uncertainty values are associated with inheritance relationships $\rightarrow$, and there are no uncertainty values or evidence counts associated with the individual terms such as $A$, $B$, etc.   This minimalism in allocation of uncertainty and attestation of evidence is a key part of the conceptual foundations of NARS.

\paragraph{ Induction}  Inductive inference in NARS has a unique and interesting semantics, similar to yet different from statistical perspectives on induction.   Operationally, given
  
\begin{itemize}
  \item $B \rightarrow A$ with $\left(f_{1}, c_{1}\right)$
  \item $B$ with $\left(f_{2}, c_{2}\right)$
\end{itemize}

\noindent one infers

\begin{itemize}
\item $A$ with $(f, c)$
\end{itemize}

\noindent via

\begin{itemize}
  \item Frequency:

$$
f=f_{1}
$$

  \item Confidence:
$$
c=\frac{f_{2} c_{2} c_{1}}{f_{2} c_{2} c_{1}+k}
$$

  \item  Power ( $f \times c$ )

$$
f \times c=\frac{f_{1} f_{2} c_{2} c_{1}}{f_{2} c_{2} c_{1}+k}
$$

\end{itemize}

The $f=f_1$ formula for conclusion frequency obviously differs drastically from anything one would obtain from Bayesian or other forms of statistical reasoning.   However, doing statistical inductive reasoning without term probabilities presents significant challenges, and because NARS eschews term probabilities it takes a different approach.

\paragraph{ Abduction}  NARS abduction bears a symmetric relationship with induction.  Given

\begin{itemize}
  \item $A \rightarrow B$ with $\left(f_{1}, c_{1}\right)$
  \item $B$ with $\left(f_{2}, c_{2}\right)$
 \end{itemize}
 
\noindent one obtains

\begin{itemize}
\item $A$ with $(f, c)$
\end{itemize}

\noindent via

\begin{itemize}
  \item Frequency:

$$
f=f_{2}
$$

\item Confidence:

$$
c=\frac{f_{1} c_{1} c_{2}}{f_{1} c_{1} c_{2}+k}
$$

  \item Power ($f \times c$ )

$$
f \times c=\frac{f_{2} f_{1} c_{1} c_{2}}{f_{1} c_{1} c_{2}+k}
$$
\end{itemize}

The same conceptual divergence with statistical methods occurs here as in the case of induction.

\subsection{Probabilistic Logic Networks (PLN)}

PLN, Probabilistic Logic Networks (Goertzel et al, 2008), was originally inspired by a desire to capture key aspects of NARS but within a framework compatible with probability theory.   The idea was, roughly, to be somewhat NARS-like in cases where evidence was scant, and then to resemble Bayesian statistical reasoning more closely in cases where evidence was more abundant.    Over time the theory and practice of PLN diverged further from NARS, and now PLN combines aspects of term logic with aspects of higher-order predicate logic, and embraces a variety of truth value representations going beyond the NARS-like $(s,c)$ = {\it (strength, confidence)} pairs with which it began.   In OpenCog Hyperon (Goertzel et al, 2023), PLN semantics is being formalized using categorial and type-theoretic methods, and connected with paraconsistent logic (Goertzel, 2020).

Here, as with NARS, we will stick to a simple subset of PLN, and focus only on the uncertain truth value formulas associated with the three first-order term logic reasoning rules for deduction, induction and abduction.   Setting aside uncertain truth value formulas, the structure of these rules looks the same for PLN as for NARS.

We will deal here only with what in PLN are called "simple truth values", which consist of strength $(s)$ and confidence $(c)$ defined via

\begin{itemize}
\item For statements:
\begin{itemize}
  \item Strength ( $s$ ): Estimated probability that a statement is true.
  \item Confidence (c): Reflects the certainty of the strength estimate.  This can be formulated in a number of ways, e.g. using imprecise probability or its variant, e.g.: The width of an $x\%$ confidence interval around the mean of a second-order distribution of which $s$ is the mean.  The more evidence that has been gathered about the statement, roughly speaking, the more peaked and less flat this second-order distribution will be.
\end{itemize}
\item For terms
\begin{itemize}
  \item Strength ( $s$ ): Estimated probability that a randomly selected object in the pertinent context, is an instance of that term
  \item Confidence (c): Reflects the certainty of the strength estimate, defined similarly to statement confidence.
\end{itemize}
\end{itemize}

A relevant subtlety here is that, in PLN, it generally doesn't make sense to do an inference without explicitly defining the inferencecontext, so that (among other uses) one can effectively estimate term probabilities.   One can of course define a "default context" associated with a given reasoning system, but this turns out to be a fairly arbitrary and not that commonly useful thing to do.   For instance, what is the term probability of the concept "hydrogen atom" versus the concept "cat"?   In a human everyday life context the term probability of "cat" is higher; in a cosmic physics context the term probability of "hydrogen atom" is higher.   In most contexts, for most human-level or more generally intelligent minds, most term probabilities will end up being assigned based on inference rather than direct observations.

NARS of course can be used to represent contexts as well; one could make terms for "everyday life context" and "cosmic physics context" and explicitly construct inheritance relations to them and manipulate these.   However it is less critical to do this in NARS than in PLN, because in NARS one doesn't have to deal with term probabilities. 

There is a fascinating connection between probabilistic mathematics and postmodern philosophy here, in that according to the latter, it is important to consider all inferences as occurring within particular contexts, for conceptual rather than technical reasons.  For our present purposes, though, we will set such depths aside and focus on truth value arithmetic!

\subsubsection{Basic PLN Truth Value Formulas}

We now review the PLN truth value formulas comparable to the NARS formulas given above.

\paragraph{Deduction}  Rather than providing single all-purpose truth value formulas corresponding to inference rules, PLN gives a methodology for deriving truth value formulas corresponding to rules.  Depending on how one wants to represent uncertainty (e.g. with first or higher order probabilities or various parametrizations or approximations thereof), and what heuristic or data-driven assumptions one wants to make, one may obtain different formulas.   For the simple (transitive inference) deduction rule, we will look here at two truth value formulas working with $(s,c)$ truth values, but founded on different assumptions otherwise.

The most standard PLN deduction truth value formula is the "independence-based" formula, which is derived using an assumption that A and C are probabilistically independent within B.   In this case, given

\begin{itemize}
  \item $A \rightarrow B$ with $s_{A B}, c_{A B}$
  \item $B \rightarrow C$ with $s_{B C}, c_{B C}$
  \item term probabilities $s_{A}, s_{B}, s_{C}$
 \end{itemize}
 
 \noindent one infers 
 
\begin{itemize}
\item $A \rightarrow C$ with $s_{A C}, c_{A C}$
\end{itemize}
 
 \noindent via

$$
s_{A C}=s_{A B} s_{B C}+\frac{\left(1-s_{A B}\right)\left(s_{C}-s_{B} s_{B C}\right)}{1-s_{B}}
$$

An alternative is the "concept geometry" based deduction formula, which replaces the independence assumption with an alternative assumption that terms have spherical (most simply, circular) shape in some feature space.   This is a crude assumption, however, the assumption underlying the independence based rule is arguably even more crude: Basically, the latter implicitly assumes terms represent random combinations of features within some feature space, with the randomness minimally constrained by any known information.  When there is data or intuitive reason to make more refined assumptions, the PLN framework can support this also.   

It should be noted that other probabilistic inference methods make similar or worse crude independence assumptions, e.g. Bayes Nets assume each node is independent of its parents conditional on its ancestors; the network discovery algorithms try to make this as true as possible, but in real cases it's always a crude approximation to reality.   NARS avoids this sort of assumption but this doesn't rule out the possibility that it implicitly makes similar assumptions -- at times its results are closer to those of the independence-based than concept-geometry-based PLN formulas, suggesting it could be cryptically relying on its own independence-like assumptions.

In the concept geometry approach, given the same premises as above,  one arrives at the formula:

$$
s_{A C}=\frac{s_{A B} s_{B C}}{\min \left(1, s_{A B}+s_{B C}\right)}
$$

\paragraph{ Induction:}  The induction rule in PLN takes inputs

\begin{itemize}
  \item $B \rightarrow A$ with $s_{B A}, c_{B A}$
  \item $B \rightarrow C$ with $s_{B C}, c_{B C}$
  \item term probabilities $s_{A}, s_{B}, s_{C}$
\end{itemize}

\noindent and infers

\begin{itemize}
\item $A \rightarrow C$ with $s_{A C}, c_{A C}$
\end{itemize}

The general approach to induction truth value formulas in PLN is to treat induction as a chaining of deduction with Bayes rule.  That is, from

\begin{itemize}
  \item $B \rightarrow A$ with $s_{B A}, c_{B A}$
\end{itemize}

\noindent one can derive

\begin{itemize}
  \item $A \rightarrow B$ with $s_{A B}, c_{A B}$
\end{itemize}

\noindent using 

$$
s_{A B} = \frac{ s_{B A} s_B}{ s_A}
$$

\noindent and then proceed with the deduction (assuming for simplicity the independence-based deduction formula)

\begin{itemize}
  \item $A \rightarrow B$ with $s_{A B}, c_{A B}$
  \item $B \rightarrow C$ with $s_{B C}, c_{B C}$
\end{itemize}

\noindent obtaining

$$
s_{A C}=s_{B A} s_{B C} \frac{s_B}{s_A} +\frac{s_B \left(1-s_{B A}\right)\left(s_{C}-s_{B} s_{B C}\right)}{ s_A (1-s_{B})}
$$

\paragraph{Abduction:} PLN abduction, symmetrically to induction, takes

\begin{itemize}
  \item $A \rightarrow B$ with $s_{A B}, c_{A B}$
  \item $C \rightarrow B$ with $s_{C B}, c_{C B}$
  \item term probabilities $s_{A}, s_{B}, s_{C}$
\end{itemize}

\noindent and infers
 
\begin{itemize}
\item  $A \rightarrow C$ with $s_{A C}, c_{A C}$
\end{itemize}

and then proceeds via Bayes' rule.

\section{Differences Between NARS and PLN As Regards Basic Truth Value Formulae}

Let's now qualitatively compare the NARS truth value formulas with the simplified, high term probability uncertainty forms of the PLN truth value formulas.

The core intuition motivating this investigation is:

\paragraph{Hypothesis} Under certain assumptions -- specifically, when term probabilities in PLN are highly uncertain -- the inference mechanisms of PLN may exhibit behaviors that resemble those of NARS.

When term probabilities $\left(s_{A}, s_{B}, s_{C}\right)$ in PLN are highly uncertain, they can be thought of as having wide variance or being imprecisely known.   This uncertainty diminishes the influence of term probabilities in PLN's inference formulas, causing the system to rely more on the strengths and confidences of the relationships between terms, resulting in the simplified formulas given above.

\paragraph{Challenges in Precise Probabilistic Calculations} One could envision a highly in-depth and rigorous approach to this sort of comparison, involving statistically studying the behavior of PLN across a variety of different configurations with different term value uncertainties.   This is a viable thing to do but seems to require either extensive numerical integration work, or extensive simulation modeling and analysis.   So we have decided to start out with the more qualitative and approximative approach as reported here.

\subsection{Heuristic Analysis for Deduction, Induction and Abduction}

For each of the three core varieties of first-order PLN and NARS inference, we want to compare $s \times c$ in PLN and $f \times c$ in NARS.

\subsubsection{Deduction}

In PLN (Independence-Based), under high uncertainty of term probabilities:

$$
s_{A C} \approx s_{A B} s_{B C}
$$

\noindent This simplification occurs because the terms involving term probabilities become less significant.

We then find:

\begin{itemize}
  \item Confidence $c_{A C} \approx c_{A B} c_{B C}$ (making a simple independence assumption on confidences).
  \item Product:

$$
s_{A C} \times c_{A C} \approx s_{A B} s_{B C} \times c_{A B} c_{B C}
$$
\end{itemize}

In PLN (Concept Geometry-Based), we have

$$
s_{A C}=\frac{s_{A B} s_{B C}}{\min \left(1, s_{A B}+s_{B C}\right)}
$$

 Under high uncertainty of term probabilities, and when $s_{A B}+s_{B C} \leq 1$, this simplifies to:

$$
s_{A C}=\frac{s_{A B} s_{B C}}{s_{A B}+s_{B C}}
$$

We have

\begin{itemize}
  \item Confidence $c_{A C} \approx c_{A B} c_{B C}$.
  \item Power:

$$
s_{A C} \times c_{A C}=\frac{s_{A B} s_{B C}}{s_{A B}+s_{B C}} \times c_{A B} c_{B C}
$$
\end{itemize}

 In NARS, comparatively, we have for strength:

$$
f=\frac{f_{1} f_{2}}{f_{1}+f_{2}-f_{1} f_{2}} \approx \frac{f_{1} f_{2}}{f_{1}+f_{2}} \quad\left(\text { for low } f_{1}, f_{2}\right)
$$

\noindent and confidence:

$$
c=c_{1} c_{2}\left(f_{1}+f_{2}-f_{1} f_{2}\right) \approx c_{1} c_{2}\left(f_{1}+f_{2}\right)
$$

Product:

$$
f \times c \approx \frac{f_{1} f_{2}}{f_{1}+f_{2}} \times c_{1} c_{2}\left(f_{1}+f_{2}\right)=f_{1} f_{2} c_{1} c_{2}
$$

\paragraph{Heuristic Comparison:}  What we can see looking at these formulas is:

\begin{itemize}
  \item Similarity: Both $s \times c$ in (the independence assumption version of) PLN and $f \times c$ in NARS involve the product of the strengths/frequencies and confidences of the premises.  That is, we literally get the same formula for the power here, though the contributions from $s / f$ versus $c$ are different in PLN vs. NARS.
  \item Difference: PLN's concept geometry-based deduction includes a normalization factor $\frac{1}{s_{A B}+s_{B C}}$, which is absent in NARS. This means that while both products involve similar multiplicative components, the scaling differs.
\end{itemize}

\subsubsection{Induction}

In the case of high uncertainty of term probabilities, the independence-based PLN induction formula

$$
s_{A C}=s_{B A} s_{B C} \frac{s_B}{s_A} +\frac{s_B \left(1-s_{B A}\right)\left(s_{C}-s_{B} s_{B C}\right)}{ s_A (1-s_{B})}
$$

\noindent can be simplified considerably.   Bayes' rule is less useful and the best approximation one can form is generally the simple form

$$
s_{A C} \approx s_{B A} s_{B C}
$$

Basically, what we are saying here is: if the term probabilities are all wildly uncertain, then heuristically speaking

\begin{itemize}
\item the best we can do as regards the first term is assume that on average the ratio of two unknown things is 1
\item since the second term has a product of two ratios of term probabilities, if term probabilities are highly uncertain then it's going to be even more wildly uncertain than the first term, and its contribution will end up negligible once confidence is incorporated
\end{itemize}

In this case PLN's default confidence formulas also reduce to a simple form

 $$
 c_{A C} \approx c_{B A} c_{B C}
 $$ 

\noindent so the power $s \times c$  comes out to:

$$
s_{A C} \times c_{A C} \approx s_{B A} s_{B C} \times c_{B A} c_{B C}
$$

In NARS, we have for strength:

$$
f=f_{1}
$$

\noindent and for confidence:

$$
c=\frac{f_{2} c_{2} c_{1}}{f_{2} c_{2} c_{1}+k}
$$

so for the power:

$$
f \times c=\frac{f_{1} f_{2} c_{2} c_{1}}{f_{2} c_{2} c_{1}+k}
$$

We see that the product depends on the frequencies and confidences of the premises, adjusted by the experience parameter $k$.

\paragraph{Heuristic Comparison:}

\begin{itemize}
  \item Similarity: Both $s_{A C} \times c_{A C}$ in PLN and $f \times c$ in NARS rely on the products of the strengths/frequencies and confidences of the premises.
  \item Difference: NARS's formula includes an additional denominator term $k$, which modulates the confidence based on the experience parameter, a concept absent in PLN.
\end{itemize}

\subsection{Abduction}

In PLN, similarly to the induction case, under high uncertainty of term probabilities we find:

$$
s_{A C} \approx s_{A B} s_{C B}
$$

\noindent yielding

\begin{itemize}
  \item Confidence $c_{A C} \approx c_{A B} c_{C B}$.
  \item Power:

$$
s_{A C} \times c_{A C} \approx s_{A B} s_{C B} \times c_{A B} c_{C B}
$$
\end{itemize}

In NARS, we have for strength

$$
f=f_{2}
$$

\noindent and for confidence:

$$
c=\frac{f_{1} c_{1} c_{2}}{f_{1} c_{1} c_{2}+k}
$$

\noindent thus the product:

$$
f \times c=\frac{f_{2} f_{1} c_{1} c_{2}}{f_{1} c_{1} c_{2}+k}
$$

Similar to induction, here the product depends on the frequencies and confidences of the premises, adjusted by the experience parameter $k$.

\paragraph{Heuristic Comparison}:  

\begin{itemize}
  \item Similarity: Both systems' products involve the multiplication of strengths/frequencies and confidences.
  \item Difference: NARS includes an experience parameter $k$ in the denominator, which affects the confidence based on accumulated experience, unlike PLN.
\end{itemize}

\section{Concrete Numerical Examples}

Following on the qualitative analysis from the previous section, to illustrate the similarities and differences between PLN and NARS under high uncertainty of term probabilities, we present three numerical examples: one for deduction, one for induction, and one for abduction. These examples demonstrate how the products $s \times c$ in PLN and $f \times c$ in NARS compare under specific conditions.

\subsection{Example 1: Deduction}
Given:

\begin{itemize}
  \item PLN:
  \begin{itemize}
  \item $A \rightarrow B$ with $s_{A B}=0.6, c_{A B}=0.8$
  \item $\quad B \rightarrow C$ with $s_{B C}=0.7, c_{B C}=0.9$
  \item High uncertainty in term probabilities
  \end{itemize}
  \item NARS:
    \begin{itemize}
  \item $A \rightarrow B$ with $f_{1}=0.6, c_{1}=0.8$
  \item $B \rightarrow C$ with $f_{2}=0.7, c_{2}=0.9$
  \item Experience parameter $k=0.5$
    \end{itemize}
\end{itemize}

\paragraph{PLN Induction (Independence-Based):}

We have a simplified $s_{A C}$ formula:

$$
s_{A C} \approx s_{A B} s_{B C}=0.6 \times 0.7=0.42
$$

\noindent and

\begin{itemize}
  \item Confidence $c_{A C} \approx c_{A B} c_{B C}=0.8 \times 0.9=0.72$
  \item Product: $s_{A C} \times c_{A C}=0.42 \times 0.72=0.3024$
\end{itemize}

\paragraph{PLN Deduction (Concept Geometry-Based):}

We have a simplified strength 

$$
s_{A C}=\frac{s_{A B} s_{B C}}{s_{A B}+s_{B C}}=\frac{0.6 \times 0.7}{0.6+0.7}=\frac{0.42}{1.3} \approx 0.3231
$$

\noindent yielding

\begin{itemize}
  \item Confidence $c_{A C} \approx c_{A B} c_{B C}=0.8 \times 0.9=0.72$
  \item Product: $s_{A C} \times c_{A C}=0.3231 \times 0.72 \approx 0.2323$
\end{itemize}

\paragraph{NARS Deduction:}

 Frequency:

$$
f=\frac{f_{1} f_{2}}{f_{1}+f_{2}-f_{1} f_{2}}=\frac{0.6 \times 0.7}{0.6+0.7-0.6 \times 0.7}=\frac{0.42}{1.3-0.42}=\frac{0.42}{0.88} \approx 0.4773
$$

Confidence:

$$
c=c_{1} c_{2}\left(f_{1}+f_{2}-f_{1} f_{2}\right)=0.8 \times 0.9 \times(0.6+0.7-0.42)=0.72 \times(1.3-0.42)=0.72 \times 0.88=0.6336
$$

 Product: 
 
 $$
 f \times c=0.4773 \times 0.6336 \approx 0.3024
 $$

\subsubsection{Analysis:}

\begin{itemize}
  \item PLN with Independence-Based Deduction:: Product $s \times c \approx 0.3024$
    \item PLN with Concept Geometry Based Deduction: Product $s \times c \approx 0.2323$
  \item NARS: Product $f \times c \approx 0.3024$
  \end{itemize}

We see that under the given simplifying assumptions, the independence-based PLN system and the NARS system yield actually identical products, indicating a strong alignment under high uncertainty of term probabilities.   

On the other hand, the normalization factor in PLN's concept geometry-based deduction results in a lower product compared to NARS.

\subsection{Example 2: Induction}

Given:

\begin{itemize}
  \item PLN:
  \begin{itemize}
  \item $B \rightarrow A$ with $s_{B A}=0.6, c_{B A}=0.8$
  \item $B \rightarrow C$ with $s_{B C}=0.7, c_{B C}=0.9$
  \item High uncertainty in term probabilities 
  \end{itemize}
  \item NARS:
  \begin{itemize}
  \item $B \rightarrow A$ with $f_{1}=0.6, c_{1}=0.8$
  \item $B$ with $f_{2}=0.7, c_{2}=0.9$
  \item Experience parameter $k=0.5$
  \end{itemize}
\end{itemize}

\paragraph{PLN Induction:} Using the simplifications introduced above we have

\begin{itemize}
  \item Simplified $s_{A C} \approx s_{B A} s_{B C}=0.6 \times 0.7=0.42$
  \item Confidence $c_{A C} \approx c_{B A} c_{B C}=0.8 \times 0.9=0.72$
  \item Product: $s_{A C} \times c_{A C}=0.42 \times 0.72=0.3024$
\end{itemize}

\paragraph{NARS Induction:}. Here we obtain

\begin{itemize}
 \item Frequency: $f=f_{1}=0.6$
  \item Confidence:
\end{itemize}

$$
c=\frac{f_{2} c_{2} c_{1}}{f_{2} c_{2} c_{1}+k}=\frac{0.7 \times 0.9 \times 0.8}{0.7 \times 0.9 \times 0.8+0.5}=\frac{0.504}{0.504+0.5}=\frac{0.504}{1.004} \approx 0.5024
$$

\begin{itemize}
  \item Product: $f \times c=0.6 \times 0.5024 \approx 0.3014$
\end{itemize}

\subsubsection{Analysis:}

\begin{itemize}
  \item (Independence based) PLN Induction:  $s \times c \approx 0.3024$
  \item NARS Induction:  $f \times c \approx 0.3014$
  \item Similarity: Both systems yield nearly identical products, indicating alignment under high uncertainty of term probabilities.
\end{itemize}

 The heuristic simplification in PLN's induction aligns closely with NARS's induction output, showcasing similar reliance on premise strengths and confidences when term probabilities are highly uncertain.

\subsection{Example 3: Abduction}

Given:

\begin{itemize}
  \item PLN:
  \begin{itemize}
  \item $A \rightarrow B$ with $s_{A B}=0.5, c_{A B}=0.7$
  \item $C \rightarrow B$ with $s_{C B}=0.4, c_{C B}=0.6$
  \item High uncertainty in term probabilities 
  \end{itemize}
  \item NARS:
  \begin{itemize}
  \item $A \rightarrow B$ with $f_{1}=0.5, c_{1}=0.7$
  \item $B$ with $f_{2}=0.4, c_{2}=0.6$
  \item Experience parameter $k=0.5$
  \end{itemize}
\end{itemize}

\paragraph{PLN Abduction:}

\begin{itemize}
  \item Simplified $s_{A C} \approx s_{A B} s_{C B}=0.5 \times 0.4=0.2$
  \item Confidence $c_{A C} \approx c_{A B} c_{C B}=0.7 \times 0.6=0.42$
  \item Product: $s_{A C} \times c_{A C}=0.2 \times 0.42=0.084$
\end{itemize}

\paragraph{NARS Abduction:}
\begin{itemize}
  \item Frequency: $f=f_{2}=0.4$
  \item Confidence:

$$
c=\frac{f_{1} c_{1} c_{2}}{f_{1} c_{1} c_{2}+k}=\frac{0.5 \times 0.7 \times 0.6}{0.5 \times 0.7 \times 0.6+0.5}=\frac{0.21}{0.21+0.5}=\frac{0.21}{0.71} \approx 0.2958
$$

  \item Product: $f \times c=0.4 \times 0.2958 \approx 0.1183$
\end{itemize}

\subsubsection{Analysis:}
\begin{itemize}
  \item PLN Abduction: $s \times c \approx 0.084$
  \item NARS Abduction:  $f \times c \approx 0.1183$
\end{itemize}

Here we see PLN yields a lower product compared to NARS, highlighting differences in how confidence is handled.

While both systems rely on the multiplicative combination of strengths/frequencies and confidences, the absence of normalization and the differing treatment of confidence in PLN lead to discrepancies in the inferred products.

\subsection{Summary of Results from Exploratory Examples}

Specifically, what we have seen here results-wise in our handful of illustrative examples is:

\paragraph{Deduction:}
\begin{itemize}
  \item Independence-Based Deduction: Demonstrates a high degree of similarity between PLN and NARS, as the simplified $s \times c$ in PLN matches $f \times c$ in NARS.
  \item Concept Geometry-Based Deduction: Shows differences due to the normalization factor, which scales the strength differently compared to NARS.
\end{itemize}

\paragraph{Induction:} The heuristic simplification in PLN's induction aligns closely with NARS's induction output, showcasing similar reliance on premise strengths and confidences when term probabilities are highly uncertain.

\paragraph{Abduction:} While PLN and NARS both rely on the multiplicative combination of strengths/frequencies and confidences, differences emerge due to PLN's lack of normalization and NARS's inclusion of an experience parameter $k$. The numerical example showed modest discrepancies in the inferred products, highlighting inherent differences.

\section{Implications and Observations}

We have shown here, through fairly simplified heuristic analyses and corresponding elementary computations, that under conditions of high uncertainty about term probabilities, the inference outputs of PLN and NARS exhibit certain converging tendencies, particularly when considering the products $s \times c$ in PLN and $f \times c$ in NARS. 

The conceptual reason for this convergence is clear: High uncertainty in term probabilities leads PLN to rely more on the strengths and confidences of relationships.   The core aim here has been to do a little specific analysis to see how this notion pans out in practice.   

The primary insight  underlying the analysis here has simply been to focus on the power $s \times c$ , because if one focuses on $s$ alone or $c$ alone the two systems do not look so similar ... sometimes what one system puts into the strength, the other system puts into the confidence, but when one takes the product it evens out.

\subsection{Similar Though Not Identical}

Overall, despite the striking parallels, we cannot say that PLN quite "reduces to" NARS when term probabilities are highly uncertain.  While strong similarities between the two sets of uncertain inference rules exist under high uncertainty of term probabilities, fundamental distinctions between the frameworks -- such as normalization factors in PLN and the experience parameter in NARS -- do lead to nontrivial differences in their reasoning outputs  even in these circumstances, especially regarding non-deductive reasoning.   These differences seem generally not huge (we have looked at more examples than the 3 given here), but modest-sized systematic differences could still have a meaningful impact on AI system behavior.   

However, it is still quite interesting how, when term probabilities are uncertain, these two inference approaches with very different conceptual and formal underpinnings come quite close to saying the same thing about practical situations -- at least if one focuses on the "power" obtained by multiplying $s \times c$.   This may be taken as a bit of evidence that $s \times c$ is an interesting thing to look at, as well.

The divergence of the results from the PLN concept-geometry based deduction formula is also interesting and perhaps merits further thought and study.   One hypothesis would be that what the concept geometry approach is attempting to do simply has no cognate in the NARS world.   Perhaps the insights and results-biasing obtained if one assumes the concepts involved in an inference have a certain natural shape in concept space, are not reflected in the NARS rules and the axioms underlying them.  On the other hand, it might be that this sort of assumption about concept shape could be reflected in NARS reasoning in some other way, perhaps at the level of the overall knowledge base rather than individual inference rules.

\subsection{The Case of Confident Term Probabilities}

When term probabilities are confidently known, the situation is generally quite different, and the role of term probabilities in PLN often leads it to very divergent conclusions from NARS.   This is very much the case for induction and abduction, where PLN's approach leans heavily on Bayes rule, which normalizes by term probabilities in a foundational and impactful way.  

In the case of reasonably confidently known term probabilities, the outputs of NARS and PLN are generally extremely different on the level of individual inference rules, and if one were to interestingly compare the two systems, it would need to be done at the level of the holistic conclusions drawn by inference networks based on a large amount of data and the combination of a large number of inferences (because in these cases, the two systems draw different conclusions on the individual-rule level, but also combine the outcomes of these individual inferences in different ways).

A more in-depth analysis could explore how rapidly and with what specificities PLN's conclusions become decreasingly NARS-ian as the confidence of term probabilities increases.

\subsection{Future Work}

To further pursue the research direction indicated here, natural avenues would be:

\begin{itemize}
  \item {\bf Numerical Simulations:} Performing intensive numerical simulations to provide more detailed quantitative comparisons, particularly in scenarios with varying levels of uncertainty.
  \item {\bf Practical Results Comparison:} Model practical examples in several domains with both systems, and assess the impact of the differences that do exist between the output values of the system, in cases of highly uncertain term probabilities and otherwise
  \item {\bf Confidence Metrics:} Explore alternative methods for calculating and interpreting PLN confidence in the presence of high uncertainty about term probabilities.
\end{itemize}

\section*{References}
\begin{itemize}
  \item Wang, Pei (1994). From Inference to Intelligence: The Non-Axiomatic Reasoning System. PhD Thesis, Indiana University.
  \item Wang, Pei (2006).  Rigid Flexibility.   Springer.
  \item Goertzel, Ben et al (2009). Probabilistic Logic Networks: A Comprehensive Framework for Uncertain Inference. Springer.
  \item Goertzel, B. (2020). Paraconsistent foundations for probabilistic reasoning, programming and concept formation. arXiv preprint arXiv:2012.14474.
\item  Goertzel, B. (2023). Generative ai vs. agi: The cognitive strengths and weaknesses of modern llms. arXiv preprint arXiv:2309.10371.
\item  Goertzel, B., Bogdanov, V., Duncan, M., Duong, D., Goertzel, Z., Horlings, J., ... \& Werko, R. (2023). OpenCog Hyperon: A Framework for AGI at the Human Level and Beyond. arXiv preprint arXiv:2310.18318.
\end{itemize}

\end{document}